# Truncating Temporal Differences:
# On the Efficient Implementation of TD($\lambda$)
# for Reinforcement Learning

**Paweł Cichosz**                                          CICHOSZ@IPE.PW.EDU.PL
*Institute of Electronics Fundamentals, Warsaw University of Technology*
*Nowowiejska 15/19, 00-665 Warsaw, Poland*

## Abstract

Temporal difference (TD) methods constitute a class of methods for learning predictions in multi-step prediction problems, parameterized by a recency factor $\lambda$. Currently the most important application of these methods is to temporal credit assignment in reinforcement learning. Well known reinforcement learning algorithms, such as AHC or Q-learning, may be viewed as instances of TD learning. This paper examines the issues of the efficient and general implementation of TD($\lambda$) for arbitrary $\lambda$, for use with reinforcement learning algorithms optimizing the discounted sum of rewards. The traditional approach, based on *eligibility traces*, is argued to suffer from both inefficiency and lack of generality. The TTD (*Truncated Temporal Differences*) procedure is proposed as an alternative, that indeed only approximates TD($\lambda$), but requires very little computation per action and can be used with arbitrary function representation methods. The idea from which it is derived is fairly simple and not new, but probably unexplored so far. Encouraging experimental results are presented, suggesting that using $\lambda > 0$ with the TTD procedure allows one to obtain a significant learning speedup at essentially the same cost as usual TD(0) learning.

## 1. Introduction

Reinforcement learning (RL, e.g., Sutton, 1984; Watkins, 1989; Barto, 1992; Sutton, Barto, & Williams, 1991; Lin, 1992, 1993; Cichosz, 1994) is a machine learning paradigm that relies on evaluative training information. At each step of discrete time a learning agent observes the current *state* of its environment and executes an *action*. Then it receives a *reinforcement* value, also called a payoff or a reward (punishment), and a state transition takes place. Reinforcement values provide a relative measure of the quality of actions executed by the agent. Both state transitions and rewards may be stochastic, and the agent does not know either transition probabilities or expected reinforcement values for any state-action combinations. The objective of learning is to identify a *decision policy* (i.e., a state-action mapping) that maximizes the reinforcement values received by the agent *in the long term*. A commonly assumed formal model of a reinforcement learning task is a *Markovian decision problem* (MDP, e.g., Ross, 1983). The *Markov property* means that state transitions and reinforcement values always depend solely on the current state and the current action: there is no dependence on previous states, actions, or rewards, i.e., the state information supplied to the agent is sufficient for making optimal decisions.

All the information the agent has about the external world and its task is contained in a series of environment states and reinforcement values. It is never told what actions to execute in particular states, or what actions (if any) would be better than those which it actually performs. It must learn an optimal policy by observing the consequences of its actions. The abstract formulation and generality of the reinforcement learning paradigm make it widely applicable, especially in such domains as game-playing (Tesauro, 1992), automatic control (Sutton et al., 1991), and robotics (Lin, 1993). To formulate a particular task as a reinforcement learning task, one just has to design appropriate state and action representation, and a reinforcement mechanism specifying the goal of the task. The main limitation of RL applications is that it is by nature a trial-and-error learning method, and it is hardly applicable in domains where making errors costs much.

A commonly studied performance measure to be maximized by an RL agent is the expected total discounted sum of reinforcement:

$$\mathbf{E}\left[\sum_{t=0}^{\infty} \gamma^t r_t\right],\tag{1}$$

where $r_t$ denotes the reinforcement value received at step $t$, and $0 \leq \gamma \leq 1$ is a *discount factor*, which adjusts the relative significance of long-term rewards versus short-term ones. To maximize the sum for any positive $\gamma$, the agent must take into account the delayed consequences of its actions: reinforcement values may be received several steps after the actions that contributed to them were performed. This is referred to as learning with *delayed reinforcement* (Sutton, 1984; Watkins, 1989). Other reinforcement learning performance measures have also been considered (Heger, 1994; Schwartz, 1993; Singh, 1994), but in this work we limit ourselves exclusively to the performance measure specified by Equation 1.

The key problem that must be solved in order to learn an optimal policy under the conditions of delayed reinforcement is known as the *temporal credit assignment* problem (Sutton, 1984). It is the problem of assigning credit or blame for the overall outcomes of a learning system (i.e., long-term reinforcement values) to each of its individual actions, possibly taken several steps before the outcomes could be observed. Discussing reinforcement learning algorithms, we will concentrate on temporal credit assignment and ignore the issues of *structural credit assignment* (Sutton, 1984), the other aspect of credit assignment in RL systems.

## 1.1 Temporal Difference Methods

The temporal credit assignment problem in reinforcement learning is typically solved using algorithms based on the methods of *temporal differences* (TD). They have been introduced by Sutton (1988) as a class of methods for learning predictions in multi-step prediction problems. In such problems prediction correctness is not revealed at once, but after more than one step since the prediction was made, though some partial information relevant to its correctness is revealed at each step. This information is available and observed as the current *state* of a prediction problem, and the corresponding prediction is computed as a value of a function of states.

Consider a multi-step prediction problem where at each step it is necessary to learn a prediction of some final outcome. It could be for example predicting the outcome of a game of chess in subsequent board situations, predicting the weather on Sunday on each day of the week, or forecasting some economic indicators. The traditional approach to learning such predictions would be to wait until the outcome occurs, keeping track of all predictions computed at intermediate steps, and then, for each of them, to use the difference between the actual outcome and the predicted value as the training error. It is supervised learning, where directed training information is obtained by comparing the outcome with predictions produced at each step. Each of the predictions is modified so as to make it closer to the outcome.

Temporal difference learning makes it unnecessary to always wait for the outcome. At each step the difference between two successive predictions is used as the training error. Each prediction is modified so as to make it closer to the next one. In fact, TD is a class of methods referred to as TD($\lambda$), where $0 \leq \lambda \leq 1$ is called a *recency factor*. Using $\lambda > 0$ allows one to incorporate prediction differences from more time steps, to hopefully speed up learning.

Temporal credit assignment in reinforcement learning may be viewed as a prediction problem. The outcome to predict in each state is simply the total discounted reinforcement that will be received starting from that state and following the current policy. Such predictions can be used for modifying the policy so as to optimize the performance measure given by Equation 1. Example reinforcement learning algorithms that implement this idea, called *TD-based algorithms*, will be presented in Section 2.2.

## 1.2 Paper Overview

Much of the research concerning TD-based reinforcement learning algorithms has concentrated on the simplest TD(0) case. However, experimental results obtained with TD($\lambda > 0$) indicate that it often allows one to obtain a significant learning speedup (Sutton, 1988; Lin, 1993; Tesauro, 1992). It has been also suggested (e.g., Peng & Williams, 1994) that TD($\lambda > 0$) should perform better in non-Markovian environments than TD(0) (i.e., it should be less sensitive to the potential violations of the Markov property). It is thus important to develop efficient and general implementation techniques that would allow TD-based RL algorithms to use arbitrary $\lambda$. This has been the motivation of this work.

The remainder of this paper is organized as follows. In Section 2 a formal definition of TD methods is presented and their application to reinforcement learning is discussed. Three example RL algorithms are briefly described: AHC (Sutton, 1984), Q-learning (Watkins, 1989; Watkins & Dayan, 1992), and advantage updating (Baird, 1993). Section 3 presents the traditional approach to TD($\lambda$) implementation, based on so called eligibility traces, which is criticized for inefficiency and lack of generality. In Section 4 the analysis of the effects of the TD algorithm leads to the formulation of the TTD (*Truncated Temporal Differences*) procedure. The two remaining sections are devoted to experimental results and concluding discussion.

## 2. Definition of TD($\lambda$)

When Sutton (1988) introduced TD methods, he assumed they would use parameter estimation techniques for prediction representation. According to his original formulation, states of a prediction problem are represented by vectors of real-valued features, and corresponding predictions are computed by the use of a set of modifiable parameters (weights). Under such representation learning consists in adjusting the weights appropriately on the basis of observed state sequences and outcomes. Below we present an alternative formula-tion, adopted from Dayan (1992), that simplifies the analysis of the effects of the TD($\lambda$) algorithm. In this formulation states may be elements of an arbitrary finite state space, and predictions are values of some function of states. Transforming Sutton's original definition of TD($\lambda$) to this alternative form is straightforward.

When discussing either the generic or RL-oriented form of TD methods, we consequently ignore the issues of function representation. It is only assumed that TD predictions or functions maintained by reinforcement learning algorithms are represented by a method that allows adjusting function values using some error values, controlled by a learning rate parameter. Whenever we write that the value of an $n$-argument function $\varphi$ for arguments $p_0, p_1, \ldots, p_{n-1}$ should be updated using an error value of $\Delta$, we mean that $\varphi(p_0, p_1, \ldots, p_{n-1})$ should be moved towards $\varphi(p_0, p_1, \ldots, p_{n-1}) + \Delta$, to a degree controlled by some learning rate factor $\eta$. The general form of this abstract update operation is written as

$$update^\eta(\varphi,\ p_0, p_1, \ldots, p_{n-1},\ \Delta). \qquad (2)$$

Under this convention, a learning algorithm is defined by the rule it uses for computing error values.

## 2.1 Basic Formulation

Let $x_0, x_1, \ldots, x_{m-1}$ be a sequence of $m$ states of a multi-step prediction problem. Each state $x_t$ can be observed at time step $t$, and at step $m$, after passing the whole sequence, a real-valued outcome $z$ can be observed. The learning system is required to produce a corresponding sequence of predictions $P(x_0), P(x_1), \ldots, P(x_{m-1})$, each of which is an estimate of $z$.

Following Dayan (1992), let us define for each state $x$:

$$\chi_x(t) = \begin{cases} 1 & \text{if } x_t = x \\ 0 & \text{otherwise.} \end{cases}$$

Then the TD($\lambda$) prediction error for each state $x$ determined at step $t$ is given by:

$$\Delta_x(t) = (P(x_{t+1}) - P(x_t)) \sum_{k=0}^{t} \lambda^{t-k} \chi_x(k), \qquad (3)$$

where $0 \leq \lambda \leq 1$ and $P(x_m) = z$ by definition, and the total prediction error for state $x$ determined after the whole observed sequence accordingly is:

$$\Delta_x = \sum_{t=0}^{m-1} \Delta_x(t) = \sum_{t=0}^{m-1} \left\{ (P(x_{t+1}) - P(x_t)) \sum_{k=0}^{t} \lambda^{t-k} \chi_x(k) \right\}. \qquad (4)$$

Thus, learning at each step is driven by the difference between two temporally successive predictions. When $\lambda > 0$, the prediction difference at time $t$ affects not only $P(x_t)$, but also predictions from previous time steps, to an exponentially decaying degree.[1]

---

1. Alternatively, learning the prediction at step $t$ relies not only on the prediction difference from that step, but also on future prediction differences. This equivalent formulation will play a significant role in Section 4.

There are two possibilities of using such defined errors for learning. The first is to compute total errors $\Delta_x$ for all states $x$, by accumulating the $\Delta_x(t)$ errors computed at each time step $t$, and to use them after passing the whole state sequence to update predictions $P(x)$. It corresponds to *batch* learning mode. The second possibility, called *incremental* or *on-line* learning, often more attractive in practice, is to update predictions at each step $t$ using current error values $\Delta_x(t)$. It is then necessary to modify appropriately Equation 3, so as to take into account that predictions are changed at each step:

$$\Delta_x(t) = (P_t(x_{t+1}) - P_t(x_t)) \sum_{k=0}^{t} \lambda^{t-k} \chi_x(k), \tag{5}$$

where $P_t(x)$ designates the prediction for state $x$ available at step $t$.

Sutton (1988) proved the convergence of batch TD(0) for a linear representation, with states represented as linearly independent vectors, under the assumption that state sequences are generated by an *absorbing Markov process*.[2] Dayan (1992) extended his proof to arbitrary $\lambda$.[3]

## 2.2 TD($\lambda$) for Reinforcement Learning

So far, this paper has presented TD as a general class of prediction methods for multi-step prediction problems. The most important application of these methods, however, is to reinforcement learning. As a matter of fact, TD methods were formulated by Sutton (1988) as a generalization of techniques he had previously used only in the context of temporal credit assignment in reinforcement learning (Sutton, 1984).

As already stated above, the most straightforward way to formulate temporal credit assignment as a prediction problem is to predict at each time step $t$ the discounted sum of future reinforcement

$$z_t = \sum_{k=0}^{\infty} \gamma^k r_{t+k},$$

called the *TD return* for time $t$. The corresponding prediction is designated by $U(x_t)$ and called the *predicted utility* of state $x_t$. TD returns obviously depend on the policy being followed; we therefore assume that $U$ values represent predicted state utilities with respect to the current policy. For perfectly accurate predictions we would have:

$$U(x_t) = z_t = r_t + \gamma z_{t+1} = r_t + \gamma U(x_{t+1}).$$

Thus, for inaccurate predictions, the mismatch or *TD error* is $r_t + \gamma U(x_{t+1}) - U(x_t)$. The resulting RL-oriented TD($\lambda$) equations take form:

$$\Delta_x(t) = (r_t + \gamma U_t(x_{t+1}) - U_t(x_t)) \sum_{k=0}^{t} (\gamma\lambda)^{t-k} \chi_x(k) \tag{6}$$

---

2. An absorbing Markov process is defined by a set of terminal states $X_T$, a set of non-terminal states $X_N$, and the set of transition probabilities $P_{xy}$ for all $x \in X_N$ and $y \in X_N \cup X_T$. The absorbing property means that any cycles among non-terminal states cannot last indefinitely long, i.e., for any starting non-terminal state a terminal state will eventually be reached (all sequences eventually terminate).

3. Recently stronger theoretical results were proved by Dayan and Sejnowski (1994) and Jaakkola, Jordan, and Singh (1993).

and

$$\Delta_x = \sum_{t=0}^{\infty} \Delta_x(t) = \sum_{t=0}^{\infty} \left\{ (r_t + \gamma U_t(x_{t+1}) - U_t(x_t)) \sum_{k=0}^{t} (\gamma \lambda)^{t-k} \chi_x(k) \right\}. \tag{7}$$

Note the following additional differences between these equations and Equations 3 and 4:

- time step subscripts are used with $U$ values to emphasize on-line learning mode,

- the discount applied in the sum in Equation 6 includes $\gamma$ as well as $\lambda$ for reasons that may be unclear now, but will be made clear in Section 4.1,

- the summation in Equation 7 extends to infinity, because the predicted final outcome is not, in general, available after any finite number of steps.

TD-based reinforcement learning algorithms may be viewed as more or less direct implementations of the general rule described by Equation 6. To see this, we will consider three algorithms: well known *AHC* (Sutton, 1984) and *Q-learning* (Watkins, 1989; Watkins & Dayan, 1992), and a recent development of Baird (1993) called *advantage updating*. All the algorithms rely on learning certain real-valued functions defined over the state or state and action space of a task. The * superscript used with any of the described functions designates its optimal values (i.e., corresponding to an optimal policy). Simplified versions of the algorithms, corresponding to TD(0), will be presented and related to Equation 6. The presentation below is limited solely to function update rules — for a more elaborated description of the algorithms the reader should consult the original publications of their developers or, for AHC and Q-learning, Lin (1993) or Cichosz (1994). They are all closely related to dynamic programming methods (Barto, Sutton, & Watkins, 1990; Watkins, 1989; Baird, 1993), but these relations, though theoretically and practically important and fruitful, are not essential for the subject of this paper and will not be discussed.

### 2.2.1 The AHC Algorithm

The variation of the AHC algorithm described here is adopted from Sutton (1990). Two functions are maintained: an *evaluation function* $V$ and a *policy function* $f$. The evaluation function evaluates each environment state and is essentially the same as what was called above the $U$ function, i.e., $V(x)$ is intended to be an estimate of the discounted sum of future reinforcement values received starting from state $x$ and following the current policy. The policy function assigns to each state-action pair $(x, a)$ a real number representing the relative merit of performing action $a$ in state $x$, called the *action merit*. The actual policy is determined from action merits using some, usually stochastic, action selection mechanism, e.g., according to a Boltzmann distribution (as described in Section 5). The optimal evaluation of state $x$, $V^*(x)$, is the expected total discounted reinforcement that will be received starting from state $x$ and following an optimal policy.

Both the functions are updated at each step $t$, after executing action $a_t$ in state $x_t$, according to the following rules:

$update^{\alpha}(V,\ x_t,\ r_t + \gamma V_t(x_{t+1}) - V_t(x_t));$

$update^{\beta}(f,\ x_t, a_t,\ r_t + \gamma V_t(x_{t+1}) - V_t(x_t)).$

The update rule for the $V$-function directly corresponds to Equation 6 for $\lambda = 0$. The update rule for the policy function increases or decreases the action merit of an action depending on whether its long-term consequences appear to be better or worse than expected. We present this, a simplified form of AHC corresponding to TD(0), because this paper proposes an alternative way of using TD($\lambda > 0$) to that implemented by the original AHC algorithm presented by Sutton (1984).

### 2.2.2 The Q-Learning Algorithm

Q-learning learns a single function of states and actions, called a *Q-function*. To each state-action pair $(x, a)$ it assigns a *Q-value* or *action utility* $Q(x, a)$, which is an estimate of the discounted sum of future reinforcement values received starting from state $x$ by executing action $a$ and then following a greedy policy with respect to the current $Q$-function (i.e., performing in each state actions with maximum $Q$-values). The current policy is implicitly defined by $Q$-values. When the optimal $Q$-function is learned, then a greedy policy with respect to action utilities is an optimal policy.

The update rule for the $Q$-function is:

$update^\alpha(Q, \ x_t, a_t, \ r_t + \gamma \max_a Q_t(x_{t+1}, a) - Q_t(x_t, a_t))$.

To show its correspondence to the TD(0) version of Equation 6, we simply assume that predicted state utilities are represented by $Q$-values so that $Q_t(x_t, a_t)$ corresponds to $U_t(x_t)$ and $\max_a Q_t(x_{t+1}, a)$ corresponds to $U_t(x_{t+1})$.

### 2.2.3 The Advantage Updating Algorithm

In advantage updating two functions are maintained: an *evaluation function $V$* and an *advantage function $A$*. The evaluation function has essentially the same interpretation as its counterpart in AHC, though it is learned in a different way. The advantage function assigns to each state-action pair $(x, a)$ a real number $A(x, a)$ representing the degree to which the expected discounted sum of future reinforcement is increased by performing action $a$ in state $x$, relative to the action currently considered best in that state. The optimal action advantages are negative for all suboptimal actions and equal 0 for optimal actions, and can be related to the optimal $Q$-values by:

$$A^*(x, a) = Q^*(x, a) - \max_{a'} Q^*(x, a').$$

Similarly as action utilities, action advantages implicitly define a policy.

The evaluation and advantage functions are updated at step $t$ by applying the following rules:

$update^\alpha(A, \ x_t, a_t, \ \max_a A_t(x_t, a) - A_t(x_t, a_t) + r_t + \gamma V_t(x_{t+1}) - V_t(x_t))$;

$update^\beta(V, \ x_t, \ \frac{1}{\alpha}[\max_a A_{t+1}(x_t) - \max_a A_t(x_t)])$.

The update rule for the advantage function is somewhat more complex that the AHC or Q-learning rules, but it still contains a term that directly corresponds to the TD(0) form of Equation 6, by replacing $V$ with $U$.

Actually, what has been presented above is a simplified version of advantage updating. The original algorithm differs in two details:

- the time step duration $\Delta t$ is explicitly included in the update rules, while in this presentation we assumed $\Delta t = 1$,

- besides *learning updates*, described above, so called *normalizing updates* are performed.

## 3. Eligibility Traces

It is obvious that the direct implementation of the computation described by Equation 6 is not too tempting. It requires maintaining $\chi_x(t)$ values for each state $x$ and past time step $t$. Note, however, that one only needs to maintain the whole sums $\sum_{k=0}^{t} (\gamma\lambda)^{t-k}\chi_x(k)$ for all $x$ and only one (current) $t$, which is much easier due to a simple trick. Substituting

$$e_x(t) = \sum_{k=0}^{t} (\gamma\lambda)^{t-k}\chi_x(k),$$

we can define the following recursive update rule:

$$
\begin{aligned}
e_x(0) &= \begin{cases} 1 & \text{if } x_0 = x \\ 0 & \text{otherwise,} \end{cases} \\
e_x(t) &= \begin{cases} \gamma\lambda e_x(t-1) + 1 & \text{if } x_t = x \\ \gamma\lambda e_x(t-1) & \text{otherwise.} \end{cases}
\end{aligned}
\tag{8}
$$

The quantities $e_x(t)$ defined this way are called *activity* or *eligibility traces* (Barto, Sutton, & Anderson, 1983; Sutton, 1984; Watkins, 1989). Whenever a state is visited, its activity becomes high and then gradually decays until it is visited again. The update to the predicted utility of each state $x$ resulting from visiting state $x_t$ at time $t$ may be then written as

$$\Delta_x(t) = (r_t + \gamma U_t(x_{t+1}) - U_t(x_t))e_x(t), \tag{9}$$

which is a direct transformation of Equation 6.

This technique (with minor differences) was already used in the early works of Barto et al. (1983) and Sutton (1984), before the actual formulation of TD($\lambda$). It is especially suitable for use with parameter estimation function representation methods, such as connectionist networks. Instead of having one $e_x$ value for each state $x$ one then has one $e_i$ value for each weight $w_i$. That is how eligibility traces were actually used by Barto et al. (1983) and Sutton (1984), inspired by an earlier work of Klopf (1982). Note that in the case of the AHC algorithm, different $\lambda$ values may be used for maintaining traces used by the evaluation and policy functions.

Unfortunately, the technique of eligibility traces is not general enough to be easy to implement with an arbitrary function representation method. It is not clear, for example, how it could be used with such an important class of function approximators as memory-based (or instance-based) function approximators (Moore & Atkeson, 1992). Applied with a pure tabular representation, it has significant drawbacks. First, it requires additional memory locations, one per state. Second, and even more painful, is that it requires modifying both $U(x)$ and $e_x$ for all $x$ at each time step. This operation dominates the computational complexity of TD-based reinforcement learning algorithms, and makes using TD($\lambda > 0$) much more expensive than TD(0). The eligibility traces implementation of TD($\lambda$) is thus, for large state spaces, absolutely impractical on serial computers, unless an appropriate function approximator is used that allows updating function values and eligibility traces for many states concurrently (such as a multi-layer perceptron). But even when such an approximator is used, there are still significant computational (both memory and time) additional costs of using TD($\lambda$) for $\lambda > 0$ versus TD(0). Another drawback of this approach will be revealed in Section 4.1.

## 4. Truncating Temporal Differences

This section departs from an alternative formulation of TD($\lambda$) for reinforcement learning. Then we follow with relating the TD($\lambda$) training errors used in this alternative formulation to TD($\lambda$) returns. Finally, we propose approximating TD($\lambda$) returns with truncated TD($\lambda$) returns, and we show how they can be computed and used for on-line reinforcement learning.

### 4.1 TD Errors and TD Returns

Let us take a closer look at Equation 7. Consider the effects of experiencing a sequence of states $x_0, x_1, \ldots, x_k, \ldots$ and corresponding reinforcement values $r_0, r_1, \ldots, r_k, \ldots$. For the sake of simplicity, assume for a while that all states in the sequence are different (though it is of course impossible for finite state spaces). Applying Equation 7 to state $x_t$ under this assumption we have:

$$
\begin{aligned}
\Delta_{x_t} &= r_t + \gamma U_t(x_{t+1}) - U_t(x_t) + \\
&\quad \gamma\lambda\Big[r_{t+1} + \gamma U_{t+1}(x_{t+2}) - U_{t+1}(x_{t+1})\Big] + \\
&\quad (\gamma\lambda)^2\Big[r_{t+2} + \gamma U_{t+2}(x_{t+3}) - U_{t+2}(x_{t+2})\Big] + \ldots \\
&= \sum_{k=0}^{\infty}(\gamma\lambda)^k\Big[r_{t+k} + \gamma U_{t+k}(x_{t+k+1}) - U_{t+k}(x_{t+k})\Big].
\end{aligned}
$$

If a state occurs several times in the sequence, each visit to that state yields a similar update. This simple observation opens a way to an alternative (though equivalent) formulation of TD($\lambda$), offering novel implementation possibilities.

Let

$$
\Delta_t^0 = r_t + \gamma U_t(x_{t+1}) - U_t(x_t) \tag{10}
$$

be the *TD(0) error* at time step $t$. We define the *TD($\lambda$) error* at time $t$ using TD(0) errors as follows:

$$
\Delta_t^\lambda = \sum_{k=0}^{\infty}(\gamma\lambda)^k\Big[r_{t+k} + \gamma U_{t+k}(x_{t+k+1}) - U_{t+k}(x_{t+k})\Big] = \sum_{k=0}^{\infty}(\gamma\lambda)^k\Delta_{t+k}^0. \tag{11}
$$

Now, we can express the overall TD($\lambda$) error for state $x$, $\Delta_x$, in terms of $\Delta_t^\lambda$ errors:

$$
\Delta_x = \sum_{t=0}^{\infty}\Delta_t^\lambda\chi_x(t). \tag{12}
$$

In fact, from Equation 7 we have:

$$\Delta_x = \sum_{t=0}^{\infty} \Delta_t^0 \sum_{k=0}^{t} (\gamma\lambda)^{t-k} \chi_x(k) = \sum_{t=0}^{\infty} \sum_{k=0}^{t} (\gamma\lambda)^{t-k} \Delta_t^0 \chi_x(k). \tag{13}$$

Swapping the order of the two summations we get:

$$\Delta_x = \sum_{k=0}^{\infty} \sum_{t=k}^{\infty} (\gamma\lambda)^{t-k} \Delta_t^0 \chi_x(k). \tag{14}$$

Finally, by exchanging $k$ and $t$ with each other, we receive:

$$\Delta_x = \sum_{t=0}^{\infty} \sum_{k=t}^{\infty} (\gamma\lambda)^{k-t} \Delta_k^0 \chi_x(t) = \sum_{t=0}^{\infty} \sum_{k=0}^{\infty} (\gamma\lambda)^{k} \Delta_{t+k}^0 \chi_x(t) = \sum_{t=0}^{\infty} \Delta_t^\lambda \chi_x(t). \tag{15}$$

Note the following important difference between $\Delta_x(t)$ (Equation 6) and $\Delta_t^\lambda$: the former is computed at each time step $t$ *for all $x$* and the latter is computed at each step $t$ only for $x_t$. Accordingly, at step $t$ the error value $\Delta_x(t)$ is used for adjusting $U(x)$ *for all $x$* and $\Delta_t^\lambda$ is only used for adjusting $U(x_t)$. This is crucial for the learning procedure proposed in Section 4.2. While applying such defined $\Delta_t^\lambda$ errors on-line makes changes to predicted state utilities at individual steps clearly different than those described by Equation 6, the overall effects of experiencing the whole state sequence (i.e., the sums of all individual error values for each state) are equivalent, as shown above.

Having expressed TD($\lambda$) in terms of $\Delta_t^\lambda$ errors, we can gain more insight into its operation and the role of $\lambda$. Some definitions will be helpful. Recall that the *TD return* for time $t$ is defined as

$$z_t = \sum_{k=0}^{\infty} \gamma^k r_{t+k}.$$

The *m-step truncated TD return* (Watkins, 1989; Barto et al., 1990) is received by taking into account only the first $m$ terms of the above sum, i.e.,

$$z_t^{[m]} = \sum_{k=0}^{m-1} \gamma^k r_{t+k}.$$

Note, however, that the rejected terms $\gamma^m r_{t+m} + \gamma^{m+1} r_{t+m+1} + \ldots$ can be approximated by $\gamma^m U_{t+m-1}(x_{t+m})$. The *corrected m-step truncated TD return* (Watkins, 1989; Barto et al., 1990) is thus:

$$z_t^{(m)} = \sum_{k=0}^{m-1} \gamma^k r_{t+k} + \gamma^m U_{t+m-1}(x_{t+m}).$$

Equation 11 may be rewritten in the following form:

$$
\begin{aligned}
\Delta_t^\lambda &= \sum_{k=0}^{\infty} (\gamma\lambda)^k \Big[ r_{t+k} + \gamma(1-\lambda)U_{t+k}(x_{t+k+1}) + \gamma\lambda U_{t+k}(x_{t+k+1}) - U_{t+k}(x_{t+k}) \Big] \\
&= \sum_{k=0}^{\infty} (\gamma\lambda)^k \Big[ r_{t+k} + \gamma(1-\lambda)U_{t+k}(x_{t+k+1}) \Big] - U_t(x_t) + \\
&\quad \sum_{k=1}^{\infty} (\gamma\lambda)^k \Big[ U_{t+k-1}(x_{t+k}) - U_{t+k}(x_{t+k}) \Big].
\end{aligned}
\tag{16}
$$

Note that for $\lambda = 1$ it yields:

$$
\begin{aligned}
\Delta_t^1 &= \sum_{k=0}^{\infty} \gamma^k r_{t+k} - U_t(x_t) + \sum_{k=1}^{\infty} \gamma^k \Big[ U_{t+k-1}(x_{t+k}) - U_{t+k}(x_{t+k}) \Big] \\
&= z_t - U_t(x_t) + \sum_{k=1}^{\infty} \gamma^k \Big[ U_{t+k-1}(x_{t+k}) - U_{t+k}(x_{t+k}) \Big].
\end{aligned}
$$

If we relax for a moment our assumption about on-line learning mode and leave out time subscripts from $U$ values, the last term disappears and we simply have:

$$
\Delta_t^1 = z_t - U(x_t).
$$

Similarly for general $\lambda$, if we define the *TD($\lambda$) return* (Watkins, 1989) for time $t$ as a weighted average of corrected truncated TD returns:

$$
z_t^\lambda = (1-\lambda) \sum_{k=0}^{\infty} \lambda^k z_t^{(k+1)} = \sum_{k=0}^{\infty} (\gamma\lambda)^k \Big[ r_{t+k} + \gamma(1-\lambda) U_{t+k}(x_{t+k+1}) \Big] \tag{17}
$$

and again omit time subscripts, we will receive:

$$
\Delta_t^\lambda = z_t^\lambda - U(x_t). \tag{18}
$$

The last equation brings more light on the exact nature of the computation performed by TD($\lambda$). The error at time step $t$ is the difference between the TD($\lambda$) return for that step and the predicted utility of the current state, that is, learning with that error value will bring the predicted utility closer to the return. For $\lambda = 1$ the quantity $z_t^\lambda$ is the usual TD return for time $t$, i.e., the discounted sum of all future reinforcement values.[4] For $\lambda < 1$ the term $r_{t+k}$ is replaced by $r_{t+k} + \gamma(1-\lambda) U_{t+k}(x_{t+k+1})$, that is, the actual immediate reward is augmented with the predicted future reward.

The definition of the TD($\lambda$) return (Equation 17) may be written recursively as

$$
z_t^\lambda = r_t + \gamma \big( \lambda z_{t+1}^\lambda + (1-\lambda) U_t(x_{t+1}) \big). \tag{19}
$$

This probably best explains the role of $\lambda$ in TD($\lambda$) learning. It determines how the return used for improving predictions is obtained. When $\lambda = 1$, it is exactly the actual observed return, the discounted sum of all rewards. For $\lambda = 0$ it is the 1-step corrected truncated return, i.e., the sum of the immediate reward and the discounted predicted utility of the successor state. Using $0 < \lambda < 1$ allows to smoothly interpolate between these two extremes, relying partially on actual returns and partially on predictions.

Equation 18 holds true only for batch learning mode, but in fact TD methods have been originally formulated for batch learning. The incremental version, more practically useful, introduces an additional term. Let $D_t^\lambda$ designate that term. By comparing Equations 16 and 17 we get:

$$D_t^\lambda = \Delta_t^\lambda - (z_t^\lambda - U_t(x_t)) = \sum_{k=1}^{\infty} (\gamma\lambda)^k \Big[ U_{t+k-1}(x_{t+k}) - U_{t+k}(x_{t+k}) \Big]. \qquad (20)$$

---

4. This observation corresponds to the equivalence of "generic" TD($\lambda$) for $\lambda = 1$ to supervised learning shown by Sutton (1988). To receive such a result it was necessary to discount prediction differences with $\gamma\lambda$ instead of $\lambda$ alone in Equation 6, though Sutton presenting the RL-oriented form of TD did not make this modification.

The magnitude of this discrepancy term, and consequently its influence on the learning process, obviously depends on the learning rate value. To examine it further, suppose a learning rate $\eta$ is used when learning $U$ on the basis of $\Delta_t^\lambda$ errors. Let the corresponding learning rule be:

$$U_{t+1}(x_t) := U_t(x_t) + \eta \Delta_t^\lambda.$$

Then we have

$$
\begin{aligned}
U_{t+1}(x_t) - U_t(x_t) &= \eta(z_t^\lambda - U_t(x_t)) + \eta D_t^\lambda \\
&= \eta(z^\lambda - U_t(x_t)) + \eta \sum_{k=1}^{\infty} (\gamma\lambda)^k \Big[ U_{t+k-1}(x_{t+k}) - U_{t+k}(x_{t+k}) \Big] \\
&\leq \eta(z^\lambda - U_t(x_t)) - \eta^2 \sum_{k=1}^{\infty} (\gamma\lambda)^k \Delta_{t+k-1}^\lambda,
\end{aligned}
\qquad (21)
$$

with equality if and only if $x_{t+k} = x_{t+k-1}$ for all $k$. A similar result may be obtained for the eligibility traces implementation, with learning driven by $\Delta_x(t)$ errors defined by Equation 9. We would then have:

$$U_{t+1}(x_t) - U_t(x_t) = \eta(z^\lambda - U_t(x_t)) - \eta^2 \sum_{k=1}^{\infty} (\gamma\lambda)^k \Delta_{t+k-1}^0 e_{x_{t+k}}(t+k-1). \qquad (22)$$

This effect may be considered another drawback of the eligibility traces implementation of TD($\lambda$), apart from its inefficiency and lack of generality. Though for small learning rates the effect of $D_t^\lambda$ is negligible, it may be still harmful in some cases, especially for large $\gamma$ and $\lambda$.[5]

## 4.2 The TTD Procedure

We have shown that TD errors $\Delta_t^\lambda$ or $z_t^\lambda - U_t(x_t)$ can be used almost equivalently for TD($\lambda$) learning, yielding the same overall results as the eligibility traces implementation, which has, however, important drawbacks in practice. Nevertheless, it is impossible to use either TD($\lambda$) errors $\Delta_t^\lambda$ or TD($\lambda$) returns $z_t^\lambda$ for on-line learning, since they are not available. At step $t$ the knowledge of both $r_{t+k}$ and $x_{t+k}$ is required *for all* $k = 1, 2, \ldots$, and there is no way to implement this in practice. Recall, however, the definition of the truncated TD return. Why not define the truncated TD($\lambda$) error and the truncated TD($\lambda$) return? The appropriate definitions are:

$$\Delta_t^{\lambda,m} = \sum_{k=0}^{m-1} (\gamma\lambda)^k \Delta_{t+k}^0 \qquad (23)$$

---

5. Sutton (1984) presented the technique of eligibility traces as an implementation of the *recency* and *frequency heuristics*. In this context, the phenomenon examined above may be considered a harmful effect of the frequency heuristic. Sutton discussed an example finite-state task where this heuristic might be misleading (Sutton, 1984, page 171).

and

$$
\begin{aligned}
z_t^{\lambda,m} &= \sum_{k=0}^{m-2} (\gamma\lambda)^k \Big[ r_{t+k} + \gamma(1-\lambda) U_{t+k}(x_{t+k+1}) \Big] + (\gamma\lambda)^{m-1} \Big[ r_{t+m-1} + \gamma U_{t+m-1}(x_{t+m}) \Big] \\
&= \sum_{k=0}^{m-1} (\gamma\lambda)^k \Big[ r_{t+k} + \gamma(1-\lambda) U_{t+k}(x_{t+k+1}) \Big] + (\gamma\lambda)^{m} U_{t+m-1}(x_{t+m}).
\end{aligned} \tag{24}
$$

We call $\Delta_t^{\lambda,m}$ the $m$-step truncated TD($\lambda$) error, or simply the $TTD(\lambda,m)$ error at time step $t$, and $z_t^{\lambda,m}$ the $m$-step truncated TD($\lambda$) return, or the $TTD(\lambda,m)$ return for time $t$. Note that $z_t^{\lambda,m}$ defined by Equation 24 is corrected, i.e., it is not obtained by simply truncating Equation 17. The correction term $(\gamma\lambda)^m U_{t+m-1}(x_{t+m})$ results in multiplying the last prediction $U_{t+m-1}(x_{t+m})$ by $\gamma$ alone instead of $\gamma(1-\lambda)$, which is virtually equivalent to using $\lambda = 0$ for that step. It is done in order to include in $z_t^{\lambda,m}$ all the available information about the expected returns for further time steps $(t+m, t+m+1, \ldots)$ contained in $U_{t+m-1}(x_{t+m})$. Without this correction for large $\lambda$ this information would be almost completely lost.

So defined, $m$-step truncated TD($\lambda$) errors or returns, can be used for on-line learning by keeping track of the last $m$ visited states, and updating at each step the predicted utility of the least recent state of those $m$ states. This idea leads to what we call the *TTD Procedure (Truncated Temporal Differences)*, which can be a good approximation of TD($\lambda$) for sufficiently large $m$. The procedure is parameterized by $\lambda$ and $m$ values. An $m$-element *experience buffer* is maintained, containing records $\langle x_{t-k}, a_{t-k}, r_{t-k}, U_{t-k}(x_{t-k+1}) \rangle$ for all $k = 0, 1, \ldots, m-1$, where $t$ is the current time step. At each step $t$ by writing $x_{[k]}$, $a_{[k]}$, $r_{[k]}$, and $u_{[k]}$ we refer to the corresponding elements of the buffer, storing $x_{t-k}$, $a_{t-k}$, $r_{t-k}$, and $U_{t-k}(x_{t-k+1})$.[6] References to $U$ are not subscripted with time steps, since all of them concern the values available at the *current* time step — in a practical implementation this directly corresponds to restoring a function value from some function approximator or a look-up table. Under this notational convention, the operation of the TTD($\lambda,m$) procedure is presented in Figure 1. It uses TTD($\lambda,m$) returns for learning. An alternative version, using TTD($\lambda,m$) errors instead (based on Equation 11), is also possible and straightforward to formulate, but there is no reason to use a "weaker" version (subject to the harmful effects described by Equations 20 and 21) when a "stronger" one is available at the same cost.

At the beginning of learning, before the first $m$ steps are made, no learning can take place. During these initial steps the operation of the TTD procedure reduces to updating appropriately the contents of the experience buffer. This obvious technical detail was left out in Figure 1 for the sake of simplicity.

The TTD($\lambda,m$) return value $z$ is computed in step 5 by the repeated application of Equation 19. The computational cost of such propagating the return in time is acceptable in practice for reasonable values of $m$. For some function representation methods, such as neural networks, the overall time complexity is dominated by the costs of retrieving a function value and learning performed in steps 4 and 6, and the cost of computing $z$ is negligible. One advantage of such implementation is that it allows to use adaptive $\lambda$ values: in step 5 one can use $\lambda_k$ depending on whether $a_{[k-1]}$ was or was not a non-policy action, or

---

6. This naturally means that the buffer's indices are shifted appropriately on each time tick.

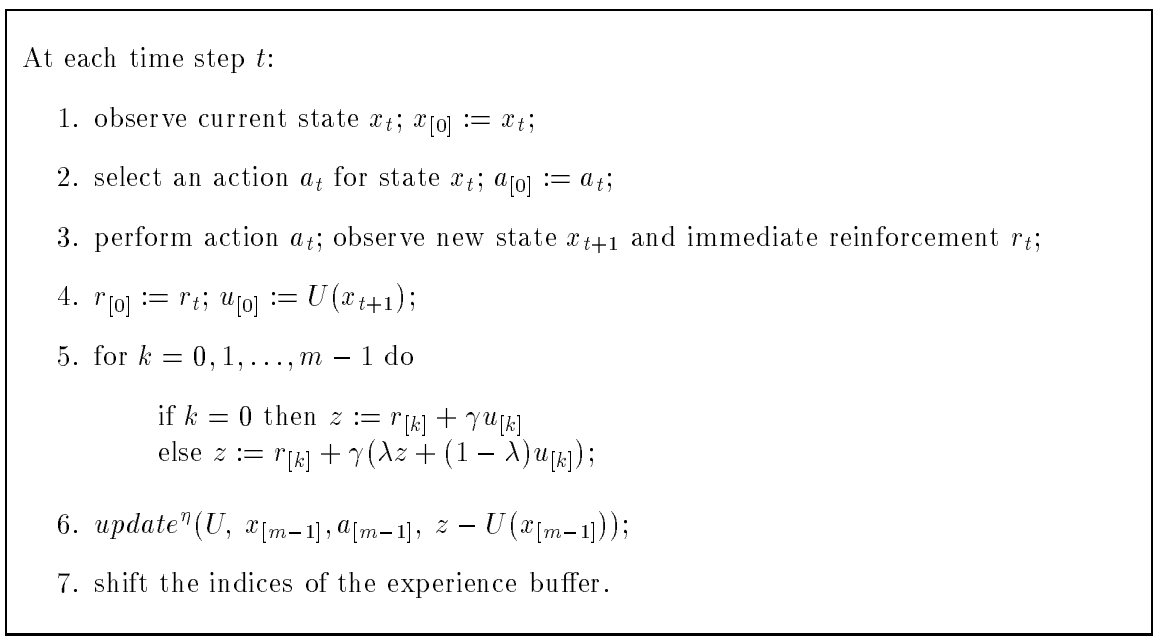

At each time step $t$:

1. observe current state $x_t$; $x_{[0]} := x_t$;

2. select an action $a_t$ for state $x_t$; $a_{[0]} := a_t$;

3. perform action $a_t$; observe new state $x_{t+1}$ and immediate reinforcement $r_t$;

4. $r_{[0]} := r_t$; $u_{[0]} := U(x_{t+1})$;

5. for $k = 0, 1, \ldots, m-1$ do

   if $k = 0$ then $z := r_{[k]} + \gamma u_{[k]}$
   else $z := r_{[k]} + \gamma(\lambda z + (1-\lambda)u_{[k]})$;

6. $update^{\eta}(U,\ x_{[m-1]}, a_{[m-1]},\ z - U(x_{[m-1]}))$;

7. shift the indices of the experience buffer.

Figure 1: The TTD($\lambda, m$) procedure.

"how much" non-policy it was. This refinement to the TD($\lambda$) algorithm was suggested by Watkins (1989) or recently Sutton and Singh (1994). Later we will see how the TTD return computation can be performed in a fully incremental way, using constant time at each step for arbitrary $m$.

Note that the function update carried out in step 6 at time $t$ applies to the state and action from time $t - m + 1$, i.e., $m - 1$ time steps earlier. This delay between an experience event and learning might be found a potential weakness of the presented approach, especially for large $m$. Note, however, that as a baseline in computing the error value the *current* utility $U(x_{[m-1]}) = U_t(x_{t-m+1})$ is used. This is an important point, because it guarantees that learning will have the desired effect of moving the utility (whatever value it currently has) towards the corresponding TTD return. If the error used in step 6 were $z - U_{t-m}(x_{t-m+1})$ instead of $z - U_t(x_{t-m+1})$, then applying it to learning at time $t$ would be problematic. Anyway, it seems that $m$ should not be too large.

The TTD procedure is not an exact implementation of TD methods for two reasons. First, it only approximates TD($\lambda$) returns with TTD($\lambda, m$) returns. Second, it introduces the aforementioned delay between experience and learning. I believe, however, that it is possible to give strict conditions under which the convergence properties of TD($\lambda$) hold true for the TTD implementation.

### 4.2.1 Choice of $m$

The reasonable choice of $m$ obviously depends on $\lambda$. For $\lambda = 0$ the best possible is $m = 1$ and for $\lambda = 1$ and $\gamma = 1$ no finite value of $m$ is large enough to accurately approximate TD($\lambda$). Fortunately, this does not seem to be very painful. It is rather unlikely that in any application one wanted to use the combination of $\lambda = 1$ and $\gamma = 1$, the more so as existing previous empirical results with TD($\lambda$) indicate that $\lambda = 1$ is usually not the optimal value to use, and it is at best comparable with other, smaller values (Sutton, 1984; Tesauro, 1992; Lin, 1993). Similar conclusions follow from the discussion of the choice of $\lambda$ presented by Watkins (1989) or Lin (1993). For $\lambda < 1$ or $\gamma < 1$ we would probably like to have such a value of $m$ that the discount $(\gamma\lambda)^m$ is a small number. One possible definition of 'small' here could be, e.g., 'much less than $\gamma\lambda$'. This is obviously a completely informal criterion. Table 1 illustrates the practical effects of this heuristic. On the other hand, for too large $m$, the delay between experience and learning introduced by the TTD procedure might become significant and cause some problems. Some of the experiments described in Section 5 have been designed in order to test different values of $m$ for fixed $0 < \lambda < 1$.

| $\gamma\lambda$ | 0.99 | 0.975 | 0.95 | 0.9 | 0.8 | 0.6 |
|---|---|---|---|---|---|---|
| $\min\{m \mid (\gamma\lambda)^m < \frac{1}{10}\gamma\lambda\}$ | 231 | 92 | 46 | 23 | 12 | 6 |

Table 1: Choosing $m$: an illustration.

### 4.2.2 Reset Operation

Until now, we have assumed that the learning process, once started, continues infinitely long. This is not true for *episodic tasks* (Sutton, 1984) and for many real-world tasks, where learning must usually stop some time. This imposes the necessity of designing a special mechanism for the TTD procedure, that will be called the *reset operation*. The reset operation would be invoked after the end of each episode in episodic tasks, or after the overall end of learning.

There is not very much to be done. The only problem that must be dealt with is that the experience buffer contains the record of the last $m$ steps for which learning has not taken place yet, and there will be no further steps that would make learning for these remaining steps possible. The implementation of the reset operation that we find the most natural and coherent with the TTD procedure is then to simulate $m$ additional fictious steps, so that learning takes place for all the real steps left in the buffer, and their TTD returns remain unaffected by the simulated fictious steps. The corresponding algorithm, presented in Figure 2, is formulated as a replacement of the original algorithm from Figure 1 for the final time step. At the final step, when there is no successor state, the fictious successor state utility is assumed to be 0. This corresponds to assigning 0 to $u_{[0]}$. The actual reset operation is performed in step 5.

### 4.2.3 Incremental TTD

As stated above, the cost of iteratively computing the TTD($\lambda, m$) return is relatively small for reasonable $m$, and with some function representation methods, for which restoring and updating function values is computationally expensive, may be really negligible. We also argued that reasonable values of $m$ should not be too large. On the other hand, such iterative return computation is easy to understand and reflects well the idea of TTD. That is why

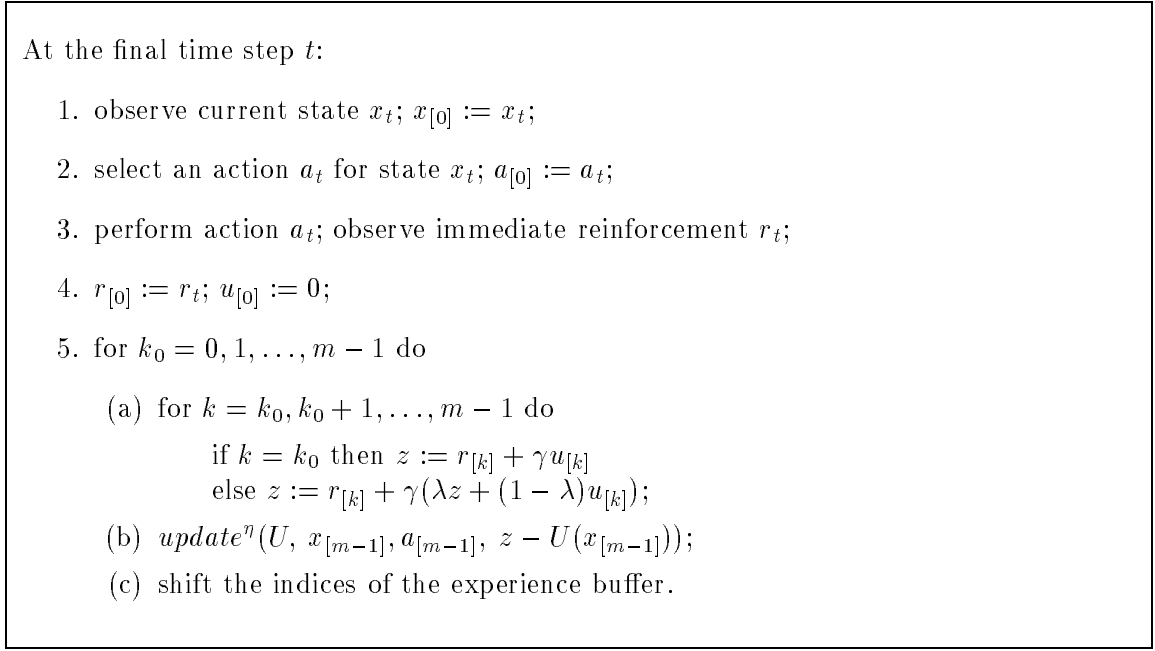

At the final time step $t$:

1. observe current state $x_t$; $x_{[0]} := x_t$;

2. select an action $a_t$ for state $x_t$; $a_{[0]} := a_t$;

3. perform action $a_t$; observe immediate reinforcement $r_t$;

4. $r_{[0]} := r_t$; $u_{[0]} := 0$;

5. for $k_0 = 0, 1, \ldots, m - 1$ do

    (a) for $k = k_0, k_0 + 1, \ldots, m - 1$ do

        if $k = k_0$ then $z := r_{[k]} + \gamma u_{[k]}$

        else $z := r_{[k]} + \gamma(\lambda z + (1 - \lambda)u_{[k]})$;

    (b) $update^\eta(U, x_{[m-1]}, a_{[m-1]}, z - U(x_{[m-1]}))$;

    (c) shift the indices of the experience buffer.

Figure 2: The reset operation for the TTD($\lambda, m$) procedure.

we presented the TTD procedure in that form. It is possible, however, to compute the TTD($\lambda, m$) return in a fully incremental manner, using constant time for arbitrary $m$.

To see this, note that the definition of the TTD($\lambda, m$) return (Equation 24) may be rewritten in the following form:

$$
\begin{aligned}
z_t^{\lambda, m} &= \sum_{k=0}^{m-1} (\gamma\lambda)^k r_{t+k} + \sum_{k=0}^{m-2} (\gamma\lambda)^k \gamma(1 - \lambda) U_{t+k}(x_{t+k+1}) + (\gamma\lambda)^{m-1} \gamma U_{t+m-1}(x_{t+m}) \\
&= S_t^{\lambda, m} + T_t^{\lambda, m} + W_t^{\lambda, m},
\end{aligned}
$$

where

$$
\begin{aligned}
S_t^{\lambda, m} &= \sum_{k=0}^{m-1} (\gamma\lambda)^k r_{t+k}, \\
T_t^{\lambda, m} &= \sum_{k=0}^{m-2} (\gamma\lambda)^k \gamma(1 - \lambda) U_{t+k}(x_{t+k+1}), \\
W_t^{\lambda, m} &= (\gamma\lambda)^{m-1} \gamma U_{t+m-1}(x_{t+m}).
\end{aligned}
$$

$W_t^{\lambda, m}$ can be directly computed in constant time for any $m$. It is not difficult to convince oneself that:

$$
S_{t+1}^{\lambda, m} = \frac{1}{\gamma\lambda} \left[ S_t^{\lambda, m} - r_t + (\gamma\lambda)^m r_{t+m} \right], \tag{25}
$$

$$
T_{t+1}^{\lambda, m} = \frac{1}{\gamma\lambda} \left[ T_t^{\lambda, m} - \gamma(1 - \lambda) U_t(x_{t+1}) + (1 - \lambda) W_t^{\lambda, m} \right]. \tag{26}
$$

The above two equations define the algorithm for computing incrementally $S_t^{\lambda,m}$ and $T_t^{\lambda,m}$, and consequently computing $z_t^{\lambda,m}$ in constant time for arbitrary $m$, with a very small computational expense. This algorithm is strictly mathematically equivalent to the algorithm presented in Figure 1.[7] Modifying appropriately the TTD procedure is straightforward and will not be discussed. A drawback of this modification is that it probably does not allow the learner to use different (adaptive) $\lambda$ values at each step, i.e., it may not be possible to combine it with the refinements suggested by Watkins (1989) or Sutton and Singh (1994). Despite this, such implementation might be beneficial if one wanted to use really large $m$.

### 4.2.4 TTD-Based Implementations of RL Algorithms

To implement particular TD-based reinforcement learning algorithms on the basis of the TTD procedure, one just has to substitute appropriate function values for $U$, and define the updating operation of step 6 in Figure 1 and step 5b in Figure 2. Specifically, for the three algorithms outlined in Section 2.2 one should:

- for AHC:

    1. replace $U(x_{t+1})$ with $V(x_{t+1})$ in step 4 (Figure 1);
    2. implement step 6 (Figure 1) and step 5b (Figure 2) as:
        $v := V(x_{[m-1]})$;
        $update^{\alpha}(V,\ x_{[m-1]},\ z-v)$;
        $update^{\beta}(f,\ x_{[m-1]}, a_{[m-1]},\ z-v)$;

- for Q-learning:

    1. replace $U(x_{t+1})$ with $\max_a Q(x_{t+1}, a)$ in step 4 (Figure 1);
    2. implement step 6 (Figure 1) and step 5b (Figure 2) as:
        $update^{\alpha}(Q,\ x_{[m-1]}, a_{[m-1]},\ z - Q(x_{[m-1]}, a_{[m-1]}))$;

- for advantage updating:

    1. replace $U(x_{t+1})$ with $V(x_{t+1})$ in step 4 (Figure 1);
    2. implement step 6 (Figure 1) and step 5b (Figure 2) as:
        $A^{\max} := \max_a A(x_{[m-1]}, a)$;
        $update^{\alpha}(A,\ x_{[m-1]}, a_{[m-1]},\ A^{\max} - A(x_{[m-1]}, a_t) + z - V(x_{[m-1]}))$;
        $update^{\beta}(V,\ x_{[m-1]},\ \frac{1}{\alpha}[\max_a A(x_{[m-1]}) - A^{\max}])$.

## 4.3 Related Work

The simple idea of truncating temporal differences that is implemented by the TTD procedure is not new. It was probably first suggested by Watkins (1989). This paper owes much to his work. But, to the best of my knowledge, this idea has never been explicitly and exactly specified, implemented, and tested. In this sense the TTD procedure is an original development.

---

7. But it is not necessarily numerically equivalent, which may sometimes cause problems in practical implementations.

Lin (1993) used a very similar implementation of TD($\lambda$), but only for what he called *experience replay*, and not for actual on-line reinforcement learning. In his approach a sequence of past experiences is replayed occasionally, and during replay for each experience the TD($\lambda$) return (truncated to the length of the replayed sequence) is computed by applying Equation 19, and a corresponding function update is performed. Such a learning method is by some means more computationally expensive than the TTD procedure (especially implemented in a fully incremental manner, as suggested above), since it requires updating predictions sequentially for all replayed experiences, besides "regular" TD(0) updates performed at each step (while TTD always requires only one update per time step), and it does not allow the learner to take full advantage of TD($\lambda > 0$), which is applied only occasionally.

Peng and Williams (1994) presented an alternative way of combining Q-learning and TD($\lambda$), different than discussed in Section 2.2. Their motivation was to better estimate TD returns by the use of TD errors. Toward that end, they used the standard Q-learning error

$$r_t + \gamma \max_a Q_t(x_{t+1}, a) - Q_t(x_t, a_t)$$

for one-step updates and a modified error

$$r_t + \gamma \max_a Q_t(x_{t+1}, a) - \max_a Q_t(x_t, a),$$

propagated using eligibility traces, thereafter. The TTD procedure achieves a similar objective in a more straightforward way, by the use of truncated TD($\lambda$) returns.

Other related work is that of Pendrith (1994). He applied the idea of eligibility traces in a non-standard way to estimate TD returns. His approach is more computationally efficient that the classical eligibility traces technique (it requires one prediction update per time step) and is free of the potentially harmful effect described by Equation 22. The method seems to be roughly equivalent to the TTD procedure with $\lambda = 1$ and large $m$, though it is probably much more implementationally complex.

## 5. Demonstrations

The demonstrations presented in this section use the AHC variant of the TTD procedure. The reason is that the AHC algorithm is the simplest of the three described algorithms and its update rule for the evaluation function most directly corresponds to TD($\lambda$). Future work will investigate the TTD procedure for the two other algorithms.

A tabular representation of the evaluation and policy functions is used. The abstract function update operation described by Equation 2 is implemented in a standard way as

$$\varphi(p_0, p_1, \ldots, p_{n-1}) := \varphi(p_0, p_1, \ldots, p_{n-1}) + \eta \Delta. \tag{27}$$

Actions to execute at each step are selected using a simple stochastic selection mechanism based on a Boltzmann distribution. According to this mechanism, action $a^*$ is selected in state $x$ with probability

$$\mathrm{Prob}(x, a^*) = \frac{\exp(f(x, a^*)/T)}{\sum_a \exp(f(x, a)/T)}, \tag{28}$$

where the *temperature* $T > 0$ adjusts the amount of randomness.

## 5.1 The Car Parking Problem

This section presents experimental results for a learning control problem with a relatively large state space and hard temporal credit assignment. We call this problem the car parking problem, though it does not attempt to simulate any real-world problem at all. Using words such as 'car', 'garage', or 'parking' is just a convention that simplifies problem description and the interpretation of results. The primary purpose of the experiments is neither just to solve the problem nor to provide evidence of the usefulness of the tested algorithm for any particular practical problem. We use this example problem in order to illustrate the performance of the AHC algorithm implemented within the TTD framework and to empirically evaluate the effects of different values of the TTD parameters $\lambda$ and $m$.

The car parking problem is illustrated in Figure 3. A car, represented as a rectangle, is initially located somewhere inside a bounded area, called the driving area. A garage is a rectangular area of a size somewhat larger than the car. All important dimensions and distances are shown in the figure. The agent — the driver of the car — is required to park it in the garage, so that the car is entirely inside. The task is episodic, though it is neither a time-until-success nor time-until-failure task (in Sutton's (1984) terminology), but rather a combination of both. Each episode finishes either when the car enters the garage or when it hits a wall (of the garage or of the driving area). After an episode the car is reset to its initial position.

### 5.1.1 State Representation

The state representation consists of three variables: the rectangular coordinates of the center of the car, $x$ and $y$, and the angle $\theta$ between the car's axis and the $x$ axis of the coordinate system. The orientation of the system is shown in the figure. The initial location and orientation of the car is fixed and described by $x = 6.15$ m, $y = 10.47$ m, and $\theta = 3.7$ rad. It was chosen so as to make the task neither too easy nor too difficult.

### 5.1.2 Action Representation

The admissible actions are 'drive straight on', 'turn left', and 'turn right'. The action of driving straight on has the effect of moving the car forward along its axis, i.e., without changing $\theta$. The actions of turning left and right are equivalent to moving along an arc with a fixed radius. The distance of each move is determined by a constant car velocity $v$ and simulation time step $\tau$. Exact motion equations and other details are given in Appendix A.

### 5.1.3 Reinforcement Mechanism

The design of the reinforcement function is fairly straightforward. The agent receives a reinforcement value of 1 (a reward) whenever it successfully parks the car in the garage, and a reinforcement value of $-1$ (a punishment) whenever it hits a wall. At all other time steps the reinforcement is 0. That is, non-zero reinforcements are received only at the last step of each episode. This involves a relatively hard temporal credit assignment problem, providing a good experimental framework for testing the efficiency of the TTD procedure. The problem is hard not only because of reinforcement delay, but also because punishments are much more frequent than rewards: it is much easier to hit a wall than to park the car correctly.

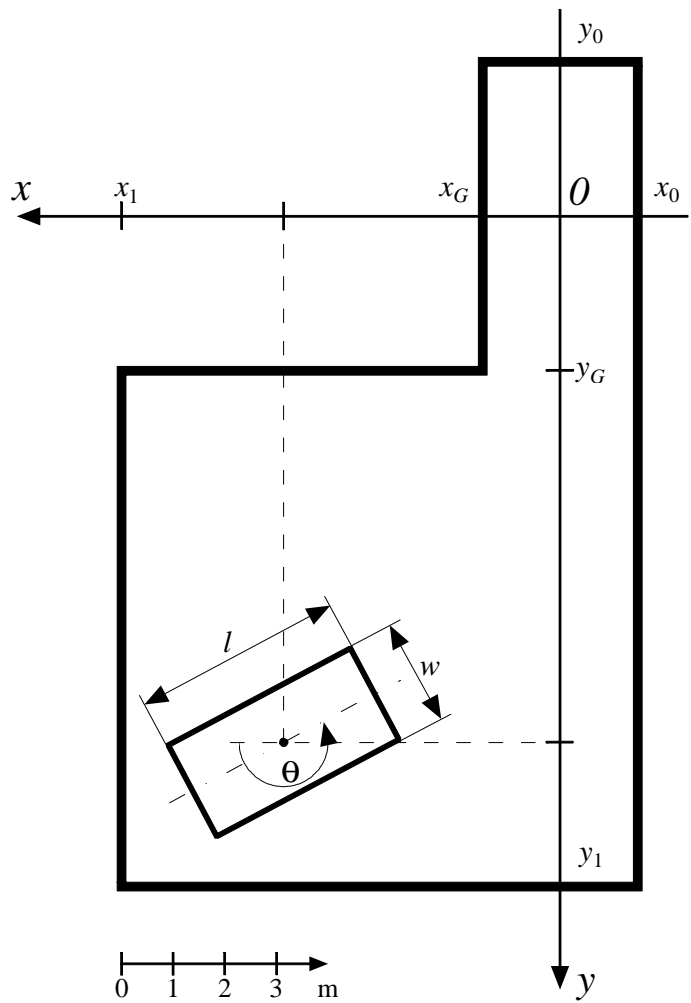

Figure 3: The car parking problem. The scale of all dimensions is preserved: $w = 2$ m, $l = 4$ m, $x_0 = -1.5$ m, $x_G = 1.5$ m, $x_1 = 8.5$ m, $y_0 = -3$ m, $y_G = 3$ m, $y_1 = 13$ m.

With such a reinforcement mechanism as presented above, an optimal policy for any $0 < \gamma < 1$ is a policy that allows to park the car in the garage in the smallest possible number of steps.

### 5.1.4 FUNCTION REPRESENTATION

The car parking problem has a continuous state space. It is artificially discretized — divided into a finite number of disjoint regions by quantizing the three state variables, and then a function value for each region is stored in a look-up table. The quantization thresholds are:

- for $x$: $-0.5, 0.0, 0.5, 1.0, 2.0, 3.0, 4.0, 6.0$ m,

- for $y$: $0.5, 1.0, 2.0, 3.0, 4.0, 5.0, 6.0, 8.0, 10.0$ m,

- for $\theta$: $\frac{19}{20}\pi, \pi, \frac{21}{20}\pi, \ldots, \frac{29}{20}\pi, \frac{3}{2}\pi, \frac{31}{20}\pi$ rad.

This yields $9 \times 10 \times 14 = 1260$ regions. Of course many of them will never be visited. The threshold values were chosen so as to make the resulting discrete state space of a moderate size. The quantization is dense near the garage, and becomes more sparse as the distance from the garage increases.

### 5.1.5 EXPERIMENTAL DESIGN AND RESULTS

Our experiments with applying the TTD procedure to the car parking problem are divided into two studies, testing the effects of the two TTD parameters $\lambda$ and $m$. The parameter settings for all experiments are presented in Table 2. The symbols $\alpha$ and $\beta$ are used to designate the learning rates for the evaluation and policy functions, respectively. The initial values of the functions were all set to 0, since we assumed that no knowledge is available about expected reinforcement levels.

| Study Number | TTD Parameters | | Learning Rates | |
|---|---|---|---|---|
| | $\lambda$ | $m$ | $\alpha$ | $\beta$ |
| 1 | 0 | 25 | 0.7 | 0.7 |
| | 0.3 | | 0.5 | 0.5 |
| | 0.5 | | 0.5 | 0.5 |
| | 0.7 | | 0.5 | 0.5 |
| | 0.8 | | 0.5 | 0.5 |
| | 0.9 | | 0.25 | 0.25 |
| | 1 | | 0.25 | 0.25 |
| 2 | 0.9 | 5 | 0.25 | 0.25 |
| | | 10 | 0.25 | 0.25 |
| | | 15 | 0.25 | 0.25 |
| | | 20 | 0.25 | 0.25 |

Table 2: Parameter settings for the experiments with the car parking problem.

As stated above, the experiments were designed to test the effects of the two TTD parameters. The other parameters were assigned values according to following principles:

- the discount factor $\gamma$ was fixed and equal 0.95 in all experiments,

- the temperature value was also fixed and set to 0.02, which seemed to be equally good for all experiments,

- the learning rates $\alpha$ and $\beta$ were roughly optimized in each experiment.[8]

Each experiment continued for 250 episodes, the number selected so as to allow all or almost all runs of all experiments to converge. The results presented for all experiments are averaged over 25 individual runs, each differing only in the initial seed of the random number generator. This number was chosen as a reasonable compromise between the reliability of results and computational costs. The results are presented as plots of the average reinforcement value per time step for the previous 5 consecutive episodes versus the episode number.

**Study 1: Effects of $\lambda$.** The objective of this study was to examine the effects of various $\lambda$ values on learning speed and quality, with $m$ set to 25. The value $m = 25$ was found to be large enough for all the tested $\lambda$ values (perhaps except $\lambda = 1$).[9] Smaller $m$ values might be used for small $\lambda$ (in particular, $m = 1$ for $\lambda = 0$), but it was kept constant for consistency.

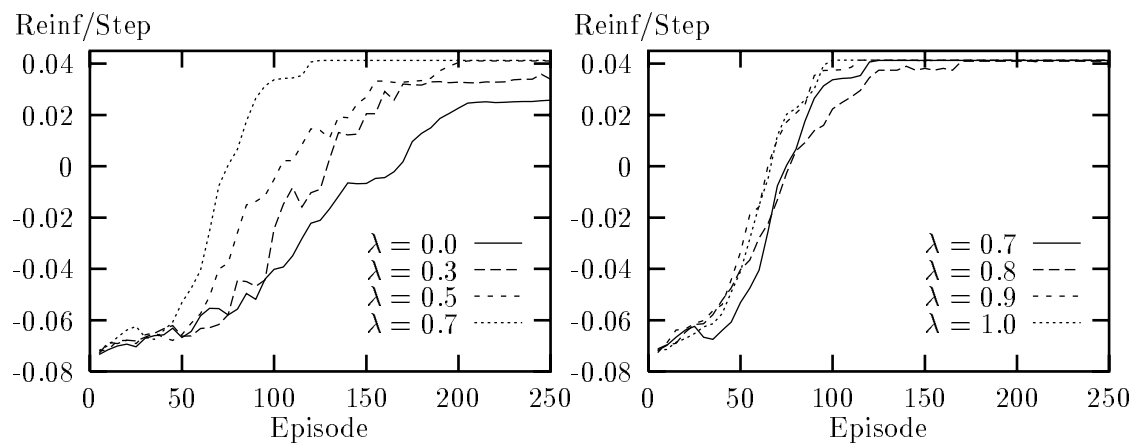

Figure 4: The car parking problem, learning curves for study 1.

The learning curves for this study are presented in Figure 4. The observations can be briefly summarized as follows:

- $\lambda = 0$ gives the worst performance of all (not all of 25 runs managed to converge within 250 episodes),

- increasing $\lambda$ improves learning speed,

- $\lambda$ values above or equal 0.7 are all similarly effective, greatly outperforming $\lambda = 0$ and clearly better than $\lambda = 0.5$,

---

8. The optimization procedure in most cases was as follows: some rather large value was tested in a few runs; if it did not give any effects of overtraining and premature convergence, it was accepted; otherwise a (usually twice) smaller value was tried, etc.

9. Note that for $\lambda = 0.9$, $m = 25$, and $\gamma = 0.95$ we have $(\gamma\lambda)^m \approx 0.02 \ll 0.855 = \gamma\lambda$.

- using large $\lambda$ caused the necessity of reducing the learning rates (cf. Table 2) to ensure convergence.

The main result is that using large $\lambda$ with the TTD procedure (including 1) always significantly improved performance. It is not quite consistent with the empirical results of Sutton (1988), who found the performance of TD($\lambda$) the best for intermediate $\lambda$, and the worst for $\lambda = 1$. Lin (1993), who used $\lambda > 0$ for his experience replay experiments, reported $\lambda$ close to 1 as the most successful, similarly as this work. He speculated that the difference between his results and Sutton's might have been caused by switching occasionally (for non-policy actions) to $\lambda = 0$ in his studies.[10] Our results, obtained for $\lambda$ held fixed all the time[11], suggest that this is not a good explanation. It seems more likely that the optimal $\lambda$ value simply strongly depends on the particular problem. Another point is that neither our TTD(1, 25) nor Lin's implementation is exactly equivalent to TD(1).

**Study 2: Effects of $m$.** This study was designed to investigate the effects of using several different $m$ values for a fixed and relatively large $\lambda$ value. The best (approximately) $\lambda$ from study 1 was used, that is 0.9. The smallest tested $m$ value is 5, which we find to be rather a small value.[12]

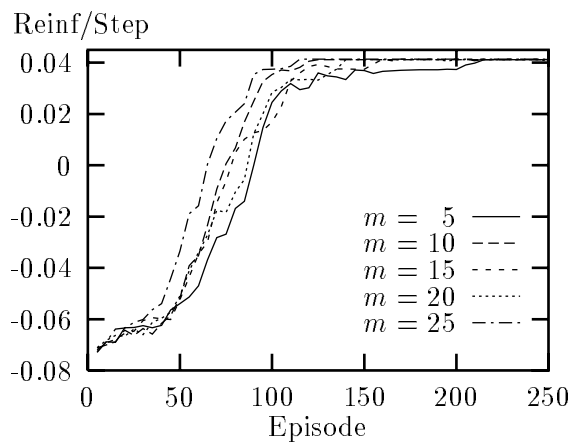

Figure 5: The car parking problem, learning curves for study 2.

The learning curves for this study are presented in Figure 5. The results for $m = 25$ were taken from study 1 for comparison. The observations can be summarized as follows:

- $m = 5$ is the worst and $m = 25$ is the best,

- the differences between intermediate $m$ values do not seem to be very statistically significant,

---

10. As a matter of fact, non-policy actions were not replayed at all in Lin's experience replay experiments.
11. Except for using $\lambda = 0$ for the most recent time step covered by the TTD return, as it follows from its definition (Equation 24).
12. For $\gamma = 0.95$, $\lambda = 0.9$, and $m = 5$ we have $(\gamma\lambda)^m \approx 0.457$, which is by all means comparable with $\gamma\lambda = 0.855$.

- even the smallest $m = 5$ gives the performance level much better than that obtained in study 1 for small $\lambda$, i.e., even relatively small $m$ values allow us to have the advantages of large $\lambda$, though larger $m$ values are generally better than small ones,

The last observation is probably the most important. It is also very optimistic. It suggests that, at least in some problems, the TTD procedure with $\lambda > 0$ allows to obtain a significant learning speed improvement over traditional TD(0)-based algorithms with practically no additional costs, because for small $m$ both space and time complexity induced by TTD is always negligible.

## 5.2 The Cart-Pole Balancing Problem

The experiments of this section have one basic purpose: to verify the effectiveness of the TTD procedure by applying its AHC implementation to a realistic and complex problem, with a long reinforcement delay, for which there exist many previous results for comparison. The cart-pole balancing problem, a classical benchmark of control specialists, is just such a problem. In particular, we would like to see whether it is possible to obtain performance (learning speed and the quality of the final policy) not worse than that reported by Barto et al. (1983) and Sutton (1984) using the eligibility traces implementation.

Figure 6 shows the cart-pole system. The cart is allowed to move along a one-dimensional bounded track. The pole can move only in the vertical plane of the cart and the track. The controller applies either a left or right force of fixed magnitude to the cart at each time step. The task is episodic: each episode finishes when a failure occurs, i.e., the pole falls or the cart hits an edge of the track. The objective is to delay the failure as long as possible.

The problem was realistically simulated by numerically solving a system of differential equations, describing the cart-pole system. These equations and other simulation details are given in Appendix B. All parameters of the simulated cart-pole system are exactly the same as used by Barto et al. (1983).

### 5.2.1 State Representation

The state of the cart-pole system is described by four state variables:

- $x$ — the position of the cart on the track,

- $\dot{x}$ — the velocity of the cart,

- $\theta$ — the angle of the pole with the vertical,

- $\dot{\theta}$ — the angular velocity of the pole.

### 5.2.2 Action Representation

At each step the agent controlling the cart-pole system chooses one of the two possible actions of applying a left or right force to the cart. The force magnitude is fixed and equal 10 N.

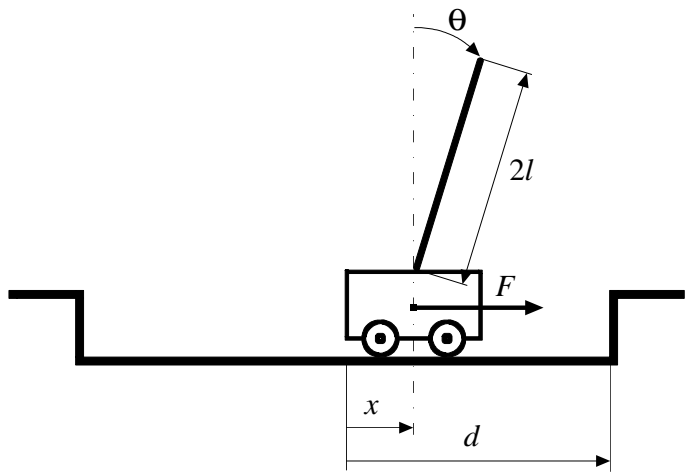

Figure 6: The cart-pole system. $F$ is the force applied to the cart's center, $l$ is a half of the pole length, and $d$ is a half of the length of the track.

### 5.2.3 REINFORCEMENT MECHANISM

The agent receives non-zero reinforcement values (namely $-1$) only at the end of each episode, i.e., after a failure. A failure occurs whenever $|\theta| > 0.21$ rad (the pole begins to fall) or $|x| > 2.4$ m (the cart hits an edge of the track). Even at the beginning of learning, with a very poor policy, an episode may continue for hundreds of time steps, and there may be many steps between a bad action and the resulting failure. This makes the temporal credit assignment problem in the cart-pole task extremely hard.

### 5.2.4 FUNCTION REPRESENTATION

As in the case of the car parking problem, we deal with the continuous state space of the cart-pole system by dividing it into disjoint regions, called boxes after Mitchie and Chambers (1968). The quantization thresholds are the same as used by Barto et al. (1983), i.e.:

- for $x$: $-0.8$, $0.8$ m,

- for $\dot{x}$: $-0.5$, $0.5$ m/s,

- for $\theta$: $-0.105$, $-0.0175$, $0$, $0.0175$, $0.105$ rad,

- for $\dot{\theta}$: $-0.8727$, $0.8727$ rad/s,

which yields $3 \times 3 \times 6 \times 3 = 162$ boxes. For each box there is a memory location, storing a function value for that box.

### 5.2.5 Experimental Design and Results

Computational expense prevented such extensive experimental studies as for the car parking problem. Only one experiment was carried out, intended to be a replication of the experiment presented by Barto et al. (1983). The values of the TTD parameters that seemed the best from the previous experiments were used, that is $\lambda = 0.9$ and $m = 25$. The discount factor $\gamma$ was set to 0.95. The learning rates for the evaluation and policy functions were roughly optimized by a small number of preliminary runs and equal $\alpha = 0.1$ and $\beta = 0.05$, respectively. The temperature of the Boltzmann distribution action selection mechanism was set to 0.0001, so as to give nearly-deterministic action selection. The initial values of the evaluation and policy functions were set to 0. We did not attempt to strictly replicate the same learning parameter values as in the work of Barto et al. (1983), since they used not only a different TD($\lambda$) implementation[13], but also a different policy representation (based on the fact that there are only two actions, while our representation is general), action selection mechanism (for the same reasons), and function learning rule.

The experiment consisted of 10 runs, differing only in the initial seed of the random number generator, and the presented results are averaged over those 10 runs. Each run continued for 100 episodes. Some of individual runs were terminated after $500,000$ time steps, before completing 100 episodes. To produce reliable averages for all 100 episodes, fictious remaining episodes were added to such runs, with the duration assigned according to the following principle, used in the experiments of Barto et al. (1983). If the duration of the last, interrupted episode was less than the duration of the immediately preceding (complete) episode, the fictious episodes were assigned the duration of that preceding episode. Otherwise, the fictious episodes were assigned the duration of the last (incomplete) episode. This prevented any short interrupted episodes from producing unreliably low averages. The results are presented in Figure 7 as plots of the average duration (the number of time steps) of the previous 5 consecutive episodes versus the episode number, in linear and logarithmic scale.

We can observe that TTD-based AHC achieved a similar (slightly better, to be exact) performance level, both as to learning speed and the quality of the final policy (i.e., the balancing periods), to that reported by Barto et al. (1983). The final balancing periods lasted above $130,000$ steps, on the average. It was obtained without using 162 additional memory locations for storing eligibility traces, and without the expensive computation necessary to update all of them at each time step, as well as all evaluation and policy function values.

## 5.3 Computational Savings

The experiments presented above illustrate the computational savings possible with the TTD procedure over conventional eligibility traces. A direct implementation of eligibility traces requires computation proportional to the number of states, i.e., to 1260 in the car parking task and to 162 in the cart-pole task — potentially many more in larger tasks. Even the straightforward iterative version of TTD may be then beneficial, as it requires computation proportional to $m$, which may be reasonably assumed to be many times less than the size of the state space. Of course, the incremental version of TTD, which requires always very small computation independent of $m$, is much more efficient.

---

13. It was the eligibility traces implementation, but eligibility traces were updated by applying a somewhat different update rule than specified by Equation 8. In particular, they were discounted with $\lambda$ alone instead of $\gamma\lambda$. Moreover, two different $\lambda$ values were used for the evaluation and policy functions.

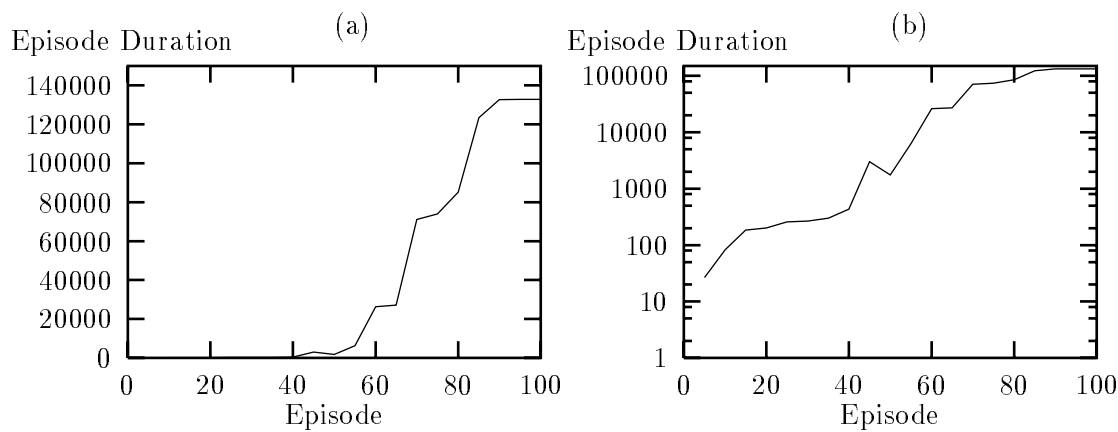

Figure 7: The cart-pole balancing problem, learning curve in (a) linear and (b) logarithmic scale.

In many practical implementations, to improve efficiency, eligibility traces and predictions are updated only for relatively few recently visited states. Traces are maintained only for the $n$ most recently visited states, and the eligibility traces of all other states are assumed to be 0.[14] But even for this "efficient" version of eligibility traces, the savings offered by TTD are considerable. For a good approximation to infinite traces in such tasks as considered here, $n$ should be at least as large as $m$. For conventional eligibility traces, there will be always a concern for keeping $n$ low, by reducing $\gamma$, $\lambda$, or the accuracy of the approximation. The same problem occurs for iterative TTD,[15] but for incremental TTD, on the other hand, none of these are at issue. The same small computation is needed independent of $m$.

## 6. Conclusion

We have informally derived the TTD procedure from the analysis of the updates introduced by TD methods to the predicted utilities of states, and shown that they can be approximated by the use of truncated TD($\lambda$) returns. Truncating temporal differences allows easy and efficient implementation. It is possible to compute TTD returns incrementally in constant time, irrespective of the value of $m$ (the truncation period), so that the computational expense of using TD-based reinforcement learning algorithms with $\lambda > 0$ is negligible (cf. Equations 25 and 26). It cannot be achieved with the eligibility traces implementation. The latter, even for such function representation methods to which it is particularly well suited (e.g., neural networks), is always associated with significant memory and time costs. The TTD procedure is probably the most computationally efficient (although most approximate) on-line implementation of TD($\lambda$). It is also general, equally good for any function representation method that might be used.

---

14. This modification cannot be applied when a parameter estimation function representation technique is used (e.g., a multi-layer perceptron), where traces are maintained for weights rather than for states.

15. The relative computational expense of iterative TTD and the "efficient" version of eligibility traces depends on the cost of the function update operation, which is always performed only for one state by the former, and for $n$ states by the latter.

An important question concerning the TTD procedure is whether its computational efficiency is not obtained at the cost of reduced learning efficiency. Having low computational costs per control action may not be attractive if the number of actions necessary to converge becomes large. As for now, no theoretically grounded answer to this important question has been provided, though it is not unlikely that such an answer will eventually be found. Nevertheless, some informal consideration may suggest that the TTD-based implementation of TD methods not only does not have to perform worse than the classical eligibility traces implementation, but it can even have some advantages. As it follows from Equations 20, 21, and 22, using TD(0) errors for on-line TD($\lambda$) learning, as in the eligibility traces implementation, introduces an additional discrepancy term, whose influence on the learning process is proportional to the square of the learning rate. That term, though often negligible, may be still harmful in certain cases, especially in tasks where the agent is likely to stay in the same states for long periods. The TTD procedure, based on truncated TD($\lambda$) returns, is free of this drawback.

Another argument supporting the TTD procedure is associated with using large $\lambda$ values, in particular 1. For an exact TD($\lambda$) implementation, such as that provided by eligibility traces, it means that learning relies solely on actually observed outcomes, without any regard to currently available predictions. It may be beneficial at the early stages of learning, when predictions are almost completely inaccurate, but in general it is rather risky — actual outcomes may be noisy and therefore sometimes misleading. The TTD procedure never relies on them entirely, even for $\lambda = 1$, since it uses $m$-step TTD returns for some finite $m$, corrected by always using $\lambda = 0$ for discounting the predicted utility of the most recent step covered by the return (cf. Equation 17). This deviation of the TTD procedure from TD($\lambda$) may turn out to be advantageous.

The TTD procedure using TTD returns for learning is only suitable for the implementation of TD methods applied to reinforcement learning. This is because in RL a part of the predicted outcome is available at each step, as the current reinforcement value. However, it is straightforward to formulate another version of the TTD procedure, using truncated TD($\lambda$) errors instead of truncated TD($\lambda$) returns, that would cover the whole scope of applications of generic TD methods.

The experimental results obtained for the TTD procedure seem very promising. The results presented in Section 5.1 show that using large $\lambda$ with the TTD procedure can give a significant performance improvement over simple TD(0) learning, even for relatively small $m$. While it does not say anything about the relative performance of TTD and the eligibility traces implementation of TD($\lambda$), it at least suggests that the TTD procedure can be useful. The best results have been obtained for the largest $\lambda$ values, including 1. This observation, contradicting to the results reported by Sutton (1988), may be a positive consequence of the TTD procedure's deviation from TD($\lambda$) discussed above.

The experiments with the cart-pole balancing problem supplied empirical evidence that for a learning control problem with a very long reinforcement delay the TTD procedure can equal or outperform the eligibility traces implementation of TD($\lambda$), even for a value of $m$ many times less than the average duration of an episode. This performance level is obtained with the TTD procedure at a much lower computational (both memory and time) expense.

To summarize, our informal consideration and empirical results suggest that the TTD procedure may have the following advantages:

- the possibility of the implementation of reinforcement learning algorithms that may be viewed as instantiations of TD($\lambda$), using $\lambda > 0$ for faster learning,

- computational efficiency: low memory requirements (for reasonable $m$) and little computation per time step,

- generality, compatibility with various function representation methods,

- good approximation of TD($\lambda$) for $\lambda < 1$ (or for $\lambda = 1$ and $\gamma < 1$),

- good practical performance, even for relatively small $m$.

There seems to be one important drawback: lack of theoretical analysis and a convergence proof. We do not know either what parameter values assure convergence or what values make it impossible. In particular, no estimate is available of the potential harmful effects of using too large $m$. Both the advantages and drawbacks cause that the TTD procedure is an interesting and promising subject for further work. This work should concentrate, on one hand, on examining the theoretical properties of this technique, and, on the other hand, on empirical studies investigating the performance of various TD-based reinforcement learning algorithms implemented within the TTD framework on a variety of problems, in particular in stochastic domains.

## Appendix A. Car Parking Problem Details

The motion of the car in the experiments of Section 5.1 is simulated by applying at each time step the following equations:

1. if $r \neq 0$ then

   (a) $\theta(t + \tau) = \theta(t) + \tau \frac{v}{r}$;

   (b) $x(t + \tau) = x(t) - r \sin \theta(t) + r \sin \theta(t + \tau)$;

   (c) $y(t + \tau) = y(t) + r \cos \theta(t) - r \sin \theta(t + \tau)$;

2. if $r = 0$ then

   (a) $\theta(t + \tau) = \theta(t)$;

   (b) $x(t + \tau) = x(t) + \tau v \cos \theta(t)$;

   (c) $y(t + \tau) = y(t) + \tau v \sin \theta(t)$;

where $r$ is the turn radius, $v$ is the car's velocity, and $\tau$ is the simulation time step. In the experiments $r = -5$ m was used for the 'turn left' action, $r = 5$ m for 'turn right', and $r = 0$ for 'drive straight on'. The velocity was constant and set to 1 m/s, and the simulation time step $\tau = 0.5$ s was used. With these parameter settings, the shortest possible path from the car's initial location ($x = 6.15$ m, $y = 10.47$ m, $\theta = 3.7$ rad) to the garage requires 21 steps.

At each step, after determining the current $x$, $y$, and $\theta$ values, the coordinates of the car's corners are computed. Then the test for intersection of each side of the car with the lines delimiting the driving area and the garage is performed to determine whether a failure occurred. If the result is negative, the test is performed for each corner of the car whether it is inside the garage, to determine if a success occurred.

## Appendix B. Cart-Pole Balancing Problem Details

The dynamics of the cart-pole system are described by the following equations of motion:

$$\ddot{x}(t) = \frac{F(t) + m_p l \left[ \dot{\theta}^2(t) \sin \theta(t) - \ddot{\theta} \cos \theta(t) \right] - \mu_c \operatorname{sgn} \dot{x}(t)}{m_c + m_p}$$

$$\ddot{\theta}(t) = \frac{g \sin \theta(t) + \cos \theta(t) \left[ \frac{-F(t) - m_p l \dot{\theta}^2(t) \sin \theta(t) + \mu_c \operatorname{sgn} \dot{x}(t)}{m_c + m_p} \right] - \frac{\mu_p \dot{\theta}(t)}{m_p l}}{l \left[ \frac{4}{3} - \frac{m_p \cos^2 \theta(t)}{m_c + m_p} \right]}$$

where

| | | | |
|---|---|---|---|
| $g$ | $=$ | $9.8$ m/s$^2$ | — acceleration due to gravity, |
| $m_c$ | $=$ | $1.0$ kg | — mass of the cart, |
| $m_p$ | $=$ | $0.1$ kg | — mass of the pole, |
| $l$ | $=$ | $0.5$ m | — half of the pole length, |
| $\mu_c$ | $=$ | $0.0005$ | — friction coefficient of the cart on the track, |
| $\mu_p$ | $=$ | $0.000002$ | — friction coefficient of the pole on the cart, |
| $F(t)$ | $=$ | $\pm 10.0$ N | — force applied to the center of the cart at time $t$. |

The equations were simulated using Euler's method with simulation time step $\tau = 0.02$ s.

## Acknowledgements

I wish to thank the anonymous reviewers of this paper for many insightful comments. I was unable to follow all their suggestions, but they contributed much to improving the paper's clarity. Thanks also to Rich Sutton, whose assistance during the preparation of the final version of this paper was invaluable.

This research was partially supported by the Polish Committee for Scientific Research under Grant 8 S503 019 05.